\documentclass[preprint,12pt]{elsarticle}

\usepackage{amssymb}

\usepackage{amsmath}
\usepackage{color}
\usepackage{multirow}

\usepackage{booktabs}

\journal{Knowledge-Based Systems}

\begin{document}

\begin{frontmatter}




\title{YOLOCS: Object Detection based on Dense Channel Compression for Feature Spatial Solidification}


\author[label1,label2,label4,label5]{Lin Huang}
\ead{h72001346@163.com}

\author[label1]{Weisheng Li \corref{cor1}}
\cortext[cor1]{Corresponding Author}
\ead{liws@cqupt.edu.cn}

\author[label2]{Yujuan Tan}
\ead{tanyujuan@cqu.edu.cn}

\author[label3]{Linlin Shen}
\ead{llshen@szu.edu.cn}

\author[label2]{Jing Yu}
\ead{20201401010@cqu.edu.cn}

\author[label1]{Haojie Fu}
\ead{1165080615@qq.com}

\affiliation[label1]{organization={Chongqing University of Posts and Telecommunications},
            addressline={No.2, Chongwen Road, Nan'an District}, 
            city={Chongqing},
            postcode={400065},
            country={China}}

\affiliation[label2]{organization={Chongqing University},
            addressline={No. 55, University Town South Road, Gaoxin District}, 
            city={Chongqing},
            postcode={401331},
            country={China}}

\affiliation[label3]{organization={Shenzhen University},
            addressline={No.3688 Nanshan Avenue, Nanshan District}, 
            city={Shenzhen},
            postcode={518061},
            country={China}}
            
\affiliation[label4]{organization={Inspur Yunzhou Industrial Internet Co., Ltd},
            addressline={No.1036 Langchao Road, Lixia District}, 
            city={Jinan},
            state={Shandong},
            postcode={250101},
            country={China}}

\affiliation[label5]{organization={Guoqi Zhimo (Chongqing) Technology Co., Ltd.},
            addressline={5th Floor, Building B15, Xiantao Data Valley, Yubei District}, 
            city={Chongqing},
            postcode={401122},
            country={China}}

\begin{abstract}

In this study, we examine the associations between channel features and convolutional kernels during the processes of feature purification and gradient backpropagation, with a focus on the forward and backward propagation within the network. Consequently, we propose a method called Dense Channel Compression  for Feature Spatial Solidification. Drawing upon the central concept of this method, we introduce two innovative modules for backbone and head networks: the Dense Channel Compression for Feature Spatial Solidification Structure (DF) and the Asymmetric Multi-Level Compression Decoupled Head (ADH). When integrated into the YOLOv5 model, these two modules demonstrate exceptional performance, resulting in a modified model referred to as YOLOCS. Evaluated on the MSCOCO dataset, the large, medium, and small YOLOCS models yield AP of 50.1\%, 47.6\%, and 42.5\%, respectively. Maintaining inference speeds remarkably similar to those of the YOLOv5 model, the large, medium, and small YOLOCS models surpass the YOLOv5 model's AP by 1.1\%, 2.3\%, and 5.2\%, respectively.

\end{abstract}




\end{frontmatter}


\section{Introduction}
YOLO\cite{yolo1}, which stands for "You Only Look Once\cite{yolo1}" is a groundbreaking real-time object detector\cite{yolo1,yolo2,yolo3,yolo4,syolo4,yolo5,yolopp,yolopp2,yolof,yolox,yolo7,focal,efficientdet,asff,ssd,kbs1,kbs2,kbs3,l2,l3,l4,l5} that has revolutionized the field of object detection. Unlike two-stage object detector\cite{maskrcnn,cascadercnn,rfh,fastrcnn,fasterrcnn,rfcn,lrcnn} that require multiple evaluations of an image through sliding windows and region proposals, YOLO performs object detection as a single regression problem. It directly predicts bounding boxes, class probabilities, and objectness scores from the entire image in one pass, making it extremely fast and efficient. Thanks to its innovative approach, YOLO has enabled real-time object detection on various platforms, including embedded systems\cite{esd}, mobile devices\cite{lmd}, and powerful GPUs\cite{gpu}. It has been widely adopted in numerous applications, such as autonomous vehicles\cite{uav}, robotics\cite{rob}, video surveillance\cite{hevs}, and augmented reality\cite{survey}, where real-time performance is crucial. Over the years, YOLO has evolved through multiple iterations, with each version improving its accuracy, speed, and performance, making it one of the most popular and efficient object detectors available today.

\begin{figure}
\centering
    \begin{minipage}[b]{0.45\linewidth}
        \centering
        \includegraphics[scale=0.45]{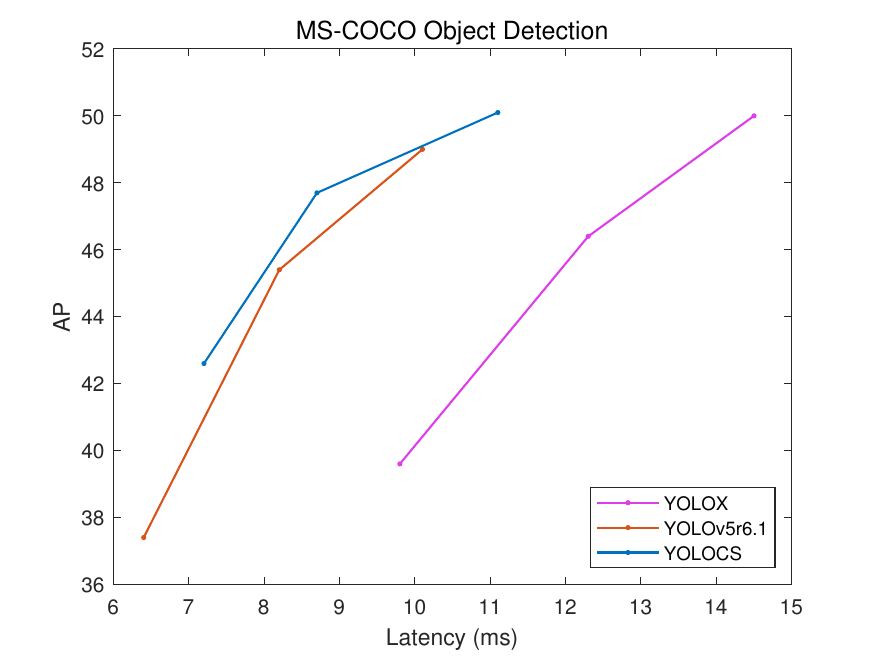}
        \centerline{(a) AP(\%) and Latency(ms, v100)}
    \end{minipage}
    \begin{minipage}[b]{0.45\linewidth}
        \centering
        \includegraphics[scale=0.45]{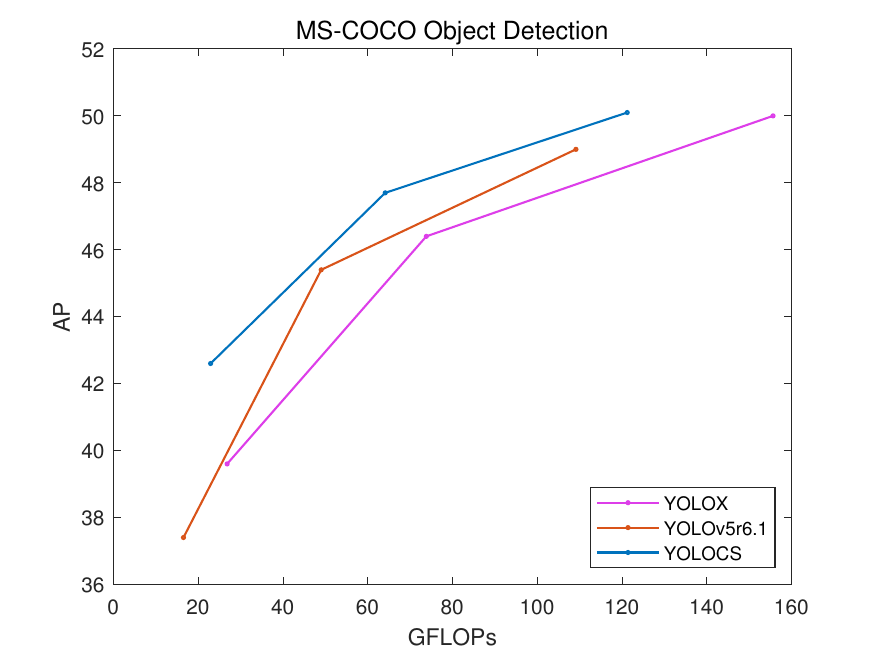}
        \centerline{(b) AP(\%) and GFLOPs}
    \end{minipage}
    \caption{Comparison of the proposed YOLOCS with YOLOv5-R6.1 and YOLOX.}
    \label{fig:data}
\end{figure}

Currently, the YOLO model can be divided into three main components: the backbone network\cite{selective}, the neck network\cite{densecross}, and the head network\cite{yolo4}. Each of these parts plays a crucial role in the overall performance of real-time object detection. In addition to the three main components of the network architecture,  the YOLO model also includes image preprocessing, Non-Maximum Suppression (NMS), loss functions, etc. Since the primary focus of this paper is to study the balance between detection accuracy and inference speed within the main network structure, these aspects are not within the scope of the discussion in this paper. The backbone network, also known as the feature extractor, is responsible for extracting meaningful features from the input image. In the backbone network, there are many innovative structures, such as Darknet-53 in YOLOv3\cite{yolo3}, Modified CSP in YOLOv4\cite{yolo4}, Modified CSP v5 in YOLOv5\cite{yolo5} and YOLOX\cite{yolox}, and Elan\cite{elan}, VoVNet\cite{vovnet}, CSPVoVNet\cite{syolo4} in YOLOv7\cite{yolo7}. These structures have played a vital role in enhancing accuracy, speed, and overall performance. However, after studying the structure of these backbone networks, we found that the trend of improving these backbone network structures is moving towards the use of fully $3 \times 3$ convolution kernels. Although $3 \times 3$ convolutions can increase the receptive field by stacking, and can also be accelerated at the GPU hardware level, the room for improvement in these structures, whether in terms of widening width or deepening depth, is still quite limited. Therefore, to address this issue, we propose a novel backbone network, namely the Dense Channel Compression for Feature Spatial Solidification Structure (DF).

The neck network serves as an bridge between the backbone and head networks, aggregating and refining the features extracted by the backbone network. It often employs structures like feature pyramid networks (FPN)\cite{fpn,l1} or spatial pyramid pooling (SPP)\cite{spp} to handle objects of various sizes and scales. The neck network is crucial for improving the detection of small and medium-sized objects, enhancing the overall accuracy of the model. The addition of the PAN\cite{pan} structure further facilitates better information flow and feature fusion, leading to improved detection performance across different object sizes. We also incorporated the Dense Channel Compression for Feature Spatial Solidification Structure (DF) from the backbone network into the neck network. However, in the neck network, this structure is similar to the CSP structure\cite{csp} in YOLOv5, without the shortcut connections\cite{resnet}. 

The head network is the final part of the YOLO model, responsible for predicting the bounding boxes, class probabilities, and objectness scores from the features provided by the neck network. In terms of innovative structures in the head network, the decoupled head in YOLOX is a prominent example. Although the decoupled head\cite{yolox} can effectively resolve the conflict between regression and classification tasks, we found through our research that the structure of the decoupled head should be consistent with the calculation logic of the loss function in order to achieve the best performance. At the same time, the performance of the decoupled head is also a focus of our attention. Therefore, to address the issues with the decoupled head, we propose a novel decoupled head structure, namely the Asymmetric Multi-level Channel Compression Decoupled Head (ADH).

\textbf{Contribution}, in this paper, YOLOv5-R6.1 is used as the baseline. The average precision (AP) of the proposed model has increased by 1.1\%, 2.3\% and 5.2\% over YOLOv5 and 0.1\%, 1.3\% and 3.0\% over YOLOX (Fig.\ref{fig:data}). The contributions of this paper can be summarized as follows:



1. We propose a novel backbone network, the Dense Channel Compression for Feature Spatial Solidification Structure (DF), which overcomes the limitations of traditional architectures. By employing dense channel compression techniques, DF solidifies the feature space in the spatial domain, optimizing the balance between network depth and width. This design ensures more effective feature extraction while preserving high inference speed, significantly enhancing overall network performance.

2. We present a novel decoupled head structure, called the Asymmetric Multi-level Channel Compression Decoupled Head (ADH). This unique design resolves the challenges associated with regression and classification tasks in existing decoupled head structures. By employing asymmetric multi-level channel compression, the ADH optimizes the consistency between the decoupled head structure and the loss function calculation logic, leading to enhanced performance and effectiveness of the decoupled head.



DF and ADH are both innovative modules presented in the paper to enhance object detection. They share the core principle of Dense Channel Compression for Feature Spatial Solidification, which focuses on optimizing feature and error propagation during forward and backward passes within a convolutional neural network. The core principle is implemented in the bottleneck structure of DF, where the number of channels in the feature map during forward propagation is progressively compressed using $3\times3$ convolutions to enhance feature extraction. Similarly, in ADH, progressive channel compression via $3\times3$ convolutions is applied to the objectness score branch network. Due to the similarity of the progressive compression structure based on $3\times3$ convolutions in both DF and ADH, the underlying principles are analogous, and they share a certain degree of correlation.

\section{Related work}
\subsection{Backbone Network}

The backbone network has consistently been the most crucial component of the YOLO series algorithms\cite{yolo1,yolo2,yolo3,yolo4,syolo4,yolo5,yolopp,yolopp2,yolof,yolox,yolo7} and the primary means of improving the accuracy of one-stage object detector. A high performance backbone network not only effectively enhances detection accuracy but also significantly reduces GFLOPs and parameters, thus boosting the model's inference speed. The backbone networks used in YOLO algorithms prior to YOLOv4 are custom-designed convolutional neural networks tailored for real-time object detection tasks. YOLOv1\cite{yolo1} features a 24-layer CNN, YOLOv2\cite{yolo2} uses the lightweight and efficient Darknet-19\cite{yolo2}, and YOLOv3 employs the more accurate Darknet-53\cite{yolo3} with residual connections\cite{resnet}. Each YOLO version prior to YOLOv4 introduced new backbone networks that improved the performance of the object detection algorithm. However, these backbone networks have limitations in training performance, inference accuracy, and inference speed due to the use of a simple stacked structure of convolutional and residual layers\cite{resnet}. Therefore, this paper focuses on the backbone networks after YOLOv4. The core idea of CSPNet\cite{csp} is widely applied to backbone networks after YOLOv4, and this paper divides its advantages into two parts: (1)The advantage in forward propagation performance, including reduced parameters, improved computational efficiency, feature fusion, and model performance; (2)The advantage in backward propagation gradient, including improved gradient flow and enhanced gradient strength. It is worth noting that the ELAN, mentioned in YOLOv7, is also based on CSPVoVNet. In summary, the improvements and innovations of the YOLOv4, YOLOv5, YOLOX, YOLOv7, and the DF structure proposed in this paper are all based on the two advantages of CSPNet applied to the backbone networks.

\subsection{Decoupled Head}

In the YOLOX, the decoupled head is a technique that separates the classification and regression branches into two separate parallel sub-networks. The classification sub-network is responsible for predicting the class label of an object, while the regression sub-network predicts the coordinates of the object's bounding box and objectness score. By decoupling these tasks, YOLOX can achieve a better trade-off between model efficiency and inference accuracy, which makes it faster and more accurate than previous YOLO models. Moreover, the decoupled head in YOLOX allows for a more efficient and flexible fine-tuning of the model on specific object categories or smaller datasets. Overall, the decoupled head in YOLOX is a powerful tool for improving the efficiency and accuracy of object detectors. This paper also optimizes and innovates the head network based on the decoupled head of YOLOX. By analyzing the computational logic between the decoupled head and the loss function, we propose a new decoupled head module called Asymmetric Multi-level Channel Compression Decoupled Head (ADH).

\section{YOLOCS}

\subsection{Dense Channel Compression for Feature Spatial Solidification Structure (DF)}

\begin{figure}
  \centering
  \includegraphics[scale=0.45]{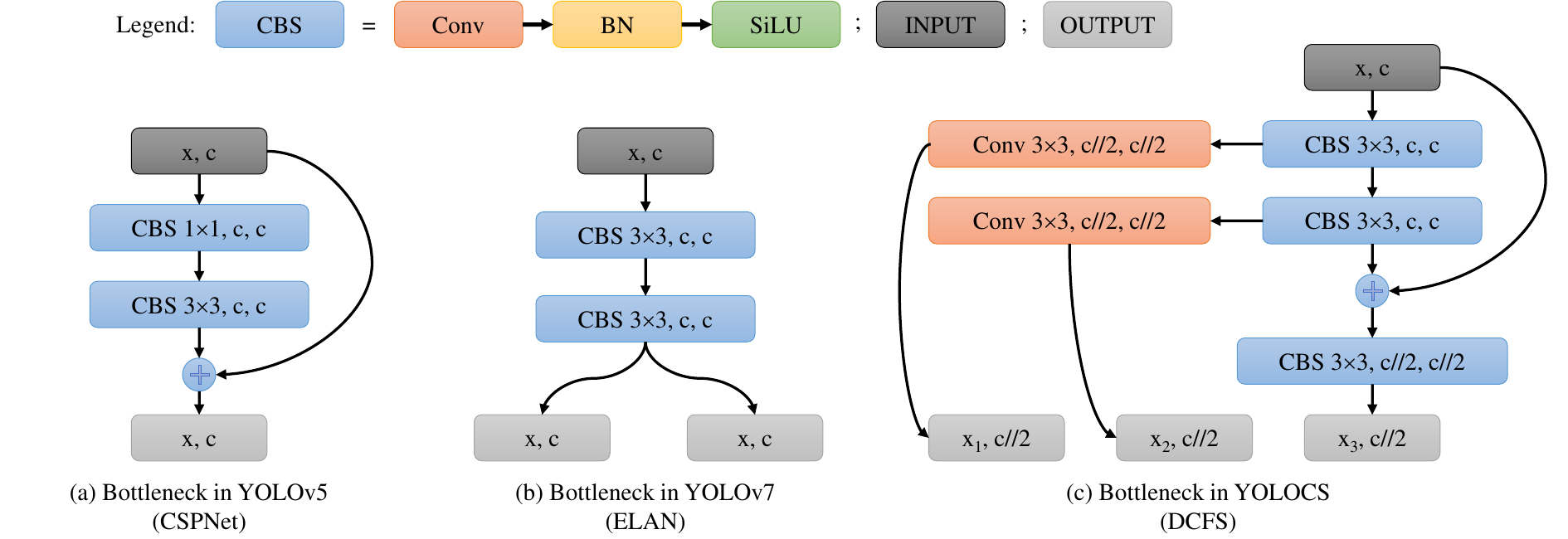}
  \caption{Bottleneck structures in backbone networks within different YOLO network architectures.}
  \label{fig:bottleneck}
\end{figure}

Within the backbone networks of the YOLO series algorithms, the most critical component, irrespective of the CSPNet structure employed in YOLOv4, YOLOv5, and YOLOX, or the ELAN structure utilized in YOLOv7, is the bottleneck structure\cite{resnet}, as depicted in Fig.\ref{fig:bottleneck}. The YOLOv5-based bottleneck structure, representative CSPNet architectures, primarily utilizes stacked residual blocks\cite{resnet} for feature extraction. The key concern with this configuration, which comprises $1\times1$ and $3\times3$ convolutions as depicted in Fig.\ref{fig:bottleneck} (a), is that even though residual connections are present, the network remains prone to overfitting as its complexity and depth escalate. To attain optimal performance, it is crucial to address not only the enhancement of network depth but also the expansion of network width, striving for a balanced and stable relationship between them\cite{efficientnet}. Nevertheless, the existing structure struggles to efficiently and adaptively extend the network width, leading to an imbalance between network depth and width that limits performance improvement. Moreover, from the perspective of backpropagation, the loss first traverses the $3\times3$ convolution and subsequently the $1\times1$ convolution. This progression can lead to the loss of features during transmission, thereby negatively impacting the efficacy of convolution kernel weight adjustments. The ELAN structure in YOLOv7 successfully expands the network width and incorporates two $3\times3$ convolutions (Fig.\ref{fig:bottleneck} (b)), which ameliorate the bottleneck structure limitations observed in YOLOv5 to a certain degree. Nevertheless, this configuration splits the output into two identical branches during forward propagation, leading to a substantial correlation between the preserved features in the current layer and those entering the next layer. As a result, the structure is unable to optimally fuse features across layers with different depths, hindering the model's performance from reaching its optimal capacity.

In our proposed method (Fig.\ref{fig:bottleneck} (c)), we not only address the balance between network width and depth, but also compress features from different depth layers through $3\times3$ convolutions, reducing the number of channels by half before outputting and fusing the features. This approach enables us to refine feature outputs from various layers to a greater extent, thereby enhancing both the diversity and effectiveness of the features at the fusion stage. Additionally, the compressed features from each layer are carried with larger convolution kernel weights ($3\times3$), thereby effectively expanding the receptive field of the output features. We term this method Dense Channel Compression for Feature Spatial Solidification. The fundamental principle behind Dense Channel Compression for Feature Spatial Solidification relies on utilizing larger convolution kernels to facilitate channel compression. This technique presents two key benefits: firstly, it expands the receptive field for feature perception during the forward propagation process, thereby ensuring that regionally correlated feature details are incorporated in order to minimize feature loss throughout the compression stage. Secondly, the enhancement of error details during the error backpropagation process allows for more accurate weight adjustments.

\begin{figure}
  \centering
  \includegraphics[scale=0.45]{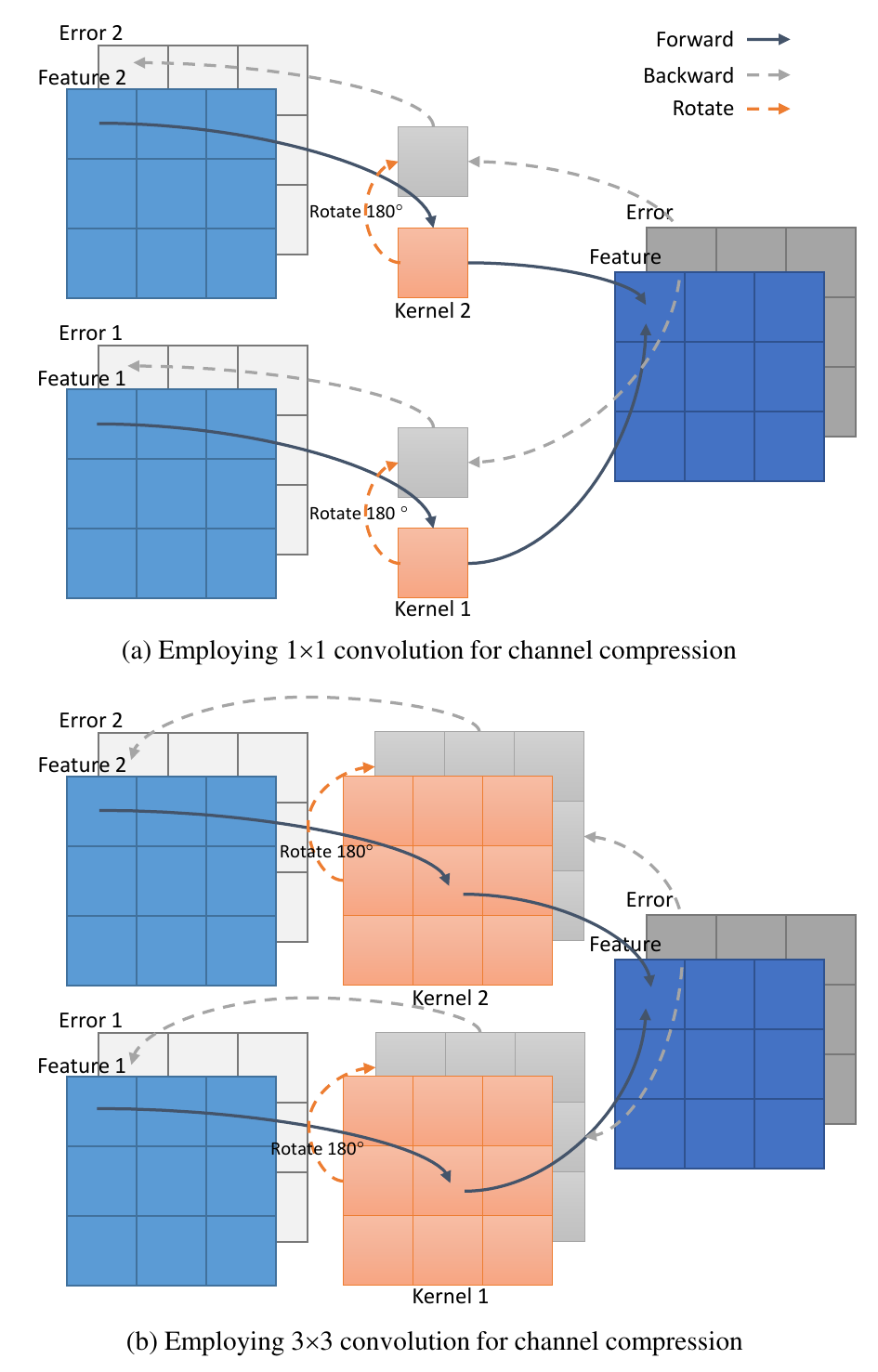}
  \caption{ Analyzing the differences in features and errors during the forward and backward propagation processes for channel compression, when utilizing convolution kernels of various sizes\cite{nc}. The orange dashed line and the label "Rotate 180$^{\circ}$" indicate that the convolution kernel used in the forward propagation is transformed into the convolution kernel used in error backpropagation by rotating it 180 degrees.}
  \label{fig:channelcomp}
\end{figure}

In order to further elucidate these two advantages, we employ convolutions with two distinct kernel types ($1\times1$ and $3\times3$) for the compression of two channels (Fig.\ref{fig:channelcomp}). The forward propagation process, represented by the black solid arrows in Fig.\ref{fig:channelcomp}, assumes input features as $X\in\mathbb{R}^{C \times H \times W}$ and output features as $Y\in\mathbb{R}^{C \times H \times W}$. The width index and height index of the feature map are denoted by $w \in \{0,1,2,...,W\}$ and $h \in \{0,1,2,...,H\}$, respectively. The computational process for channel compression, utilizing a $1\times1$ convolution (Fig.\ref{fig:channelcomp} (a)), can be summarized as follows:

\begin{equation}
  Y_{h,w} = \sum_{c=0}^C(X_{h,w,c}K_c + B_c)
\end{equation}
where $c \in \{0,1,2,...,C\}$ represents the channel index, $K$ denotes the convolutional kernel, and $B$ refers to the bias term. The computational process for channel compression, utilizing a $3\times3$ convolution (Fig.\ref{fig:channelcomp} (b)), can be summarized as follows:

\begin{equation}
\begin{aligned}
  Y_{h,w} = \sum_{c=0}^C(\sum_{k=0}^{KS}X_{h + \lfloor\frac{k}{|\sqrt{KS}|}\rfloor - \lfloor \frac{KS}{2} \rfloor, w + k\%|\sqrt{KS}| - \lfloor \frac{KS}{2} \rfloor,c}K_{c,k} + B_c)
\end{aligned}
\end{equation}
where $KS$ represents the size of the convolutional kernel (e.g. $3\times3$ convolutional kernels $KS=9$), and $k \in \{0,1,2,...,KS\}$ is the convolutional kernel index. 

By examining the two formulas, it becomes evident that for pixels ($Y_{h,w}$) located at identical positions in both input and output feature maps, employing a $3\times3$ convolution preserves the neighboring pixel features, which are also retained in the surrounding pixels. In the context of forward propagation, Equation 1 (using 1×1 convolutions) performs a summation over the channels at each spatial location, whereas Equation 2 (using 3×3 convolutions) involves a more complex summation over neighboring spatial locations, integrating information from a 3×3 region. This larger convolution kernel not only captures richer spatial context but also improves weight adjustments during backward propagation, ensuring that features from a broader area contribute to both forward and backward passes. This results in more effective feature recovery and error correction, further solidifying spatial information and boosting model performance.

\begin{figure}
  \centering
  \includegraphics[scale=0.4]{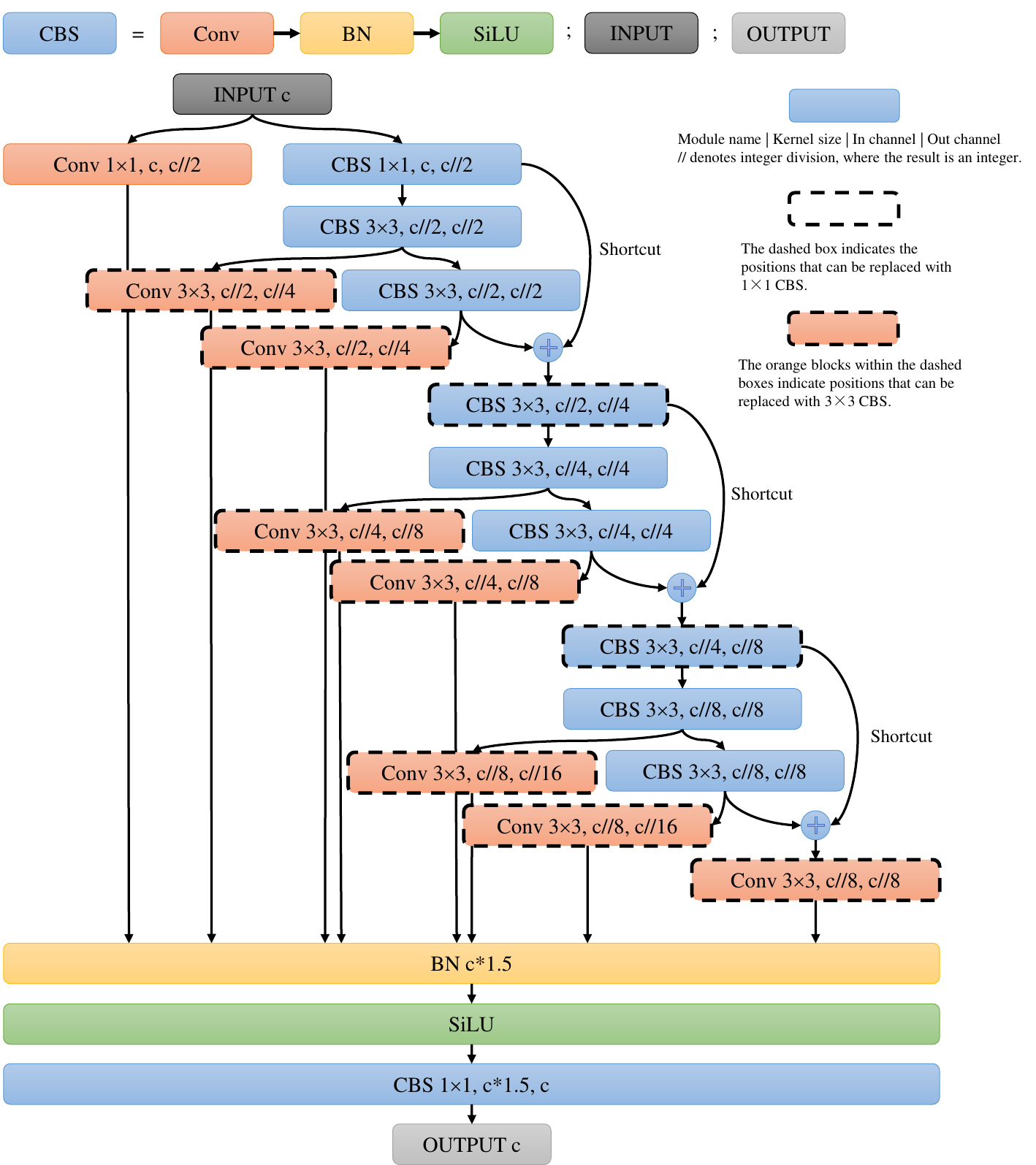}
  \caption{ The dense channel compression for feature Spatial Solidification architecture is fully incorporated into the backbone network, with the shortcut component removed from the neck structure.}
  \label{fig:dcfs}
\end{figure}

Additionally, from a backward propagation standpoint (indicated by the gray dashed arrows in Fig.\ref{fig:channelcomp}), the error during backward propagation renders channel compression equivalent to channel expansion\cite{nc}. Utilizing $1\times1$ convolution for error feature expansion across different channels has limitations, as it can only obtain information from single pixel errors. This results in a constrained receptive field incapable of acquiring critical contextual information from neighboring pixel errors, yielding suboptimal error propagation performance. Conversely, adopting a $3\times3$ convolution for error feature expansion in different channels facilitates the merging of error information from more extensive regions, thereby improving error propagation, enhancing error propagation performance, and ultimately boosting the weight adjustment effect. As a result, utilizing larger convolution kernels for channel compression allows for the optimal preservation of feature (error) details from the preceding layer in both forward and backward propagation. This approach reduces the number of channels while sustaining inference accuracy, ultimately enhancing the model's overall inference speed. This principle constitutes the core innovation of our proposed method. Moreover, the comparative experimental data of DF ($3 \times 3$, replace the orange block in Fig.\ref{fig:dcfs} with CBS) and DF ($1 \times 1$, replace the convolution of the black dashed box in Fig.\ref{fig:dcfs} with CBS) in Tab.\ref{tab:comp} further substantiates this point by showcasing a superior performance of $3\times3$ convolution over $1\times1$ convolution by 0.3\%.

\begin{table}[h]
\centering
	\caption{The AP(\%), Parameters, GFLOPs of DF($1\times1$ Conv), DF($3\times3$ Conv), and DF(OCFJ), using YOLOv5l-R6.1 as the baseline, on the MS-COCO val2017. In the table, green indicates the best AP.}
	\label{tab:comp}
	\begin{tabular}{llll}
	\toprule
    Models & AP(\%) & Params(M) & GFLOPs\\
    \midrule
    v5l & 49.0 & 46.5 & 109.1 \\
    DF($1\times1$) & 49.1 (+0.1)& 44.8 (-1.7) & 94.5 (-14.6)\\
    DF($3\times3$) & 49.4 (+0.4)& 56.9 (+10.4)& 121.3 (+12.2)\\
    DF(OCFJ) & 49.7 (\textcolor{green}{+0.7}) & 52.4 (+5.9) & 113.1 (+4.0) \\
    \bottomrule
\end{tabular}
\end{table}

The network structure of the Dense Channel Compression for Feature Spatial Solidification structure (DF) is depicted in Fig.\ref{fig:dcfs}. We employ a three-layer bottleneck structure, which gradually compresses the channels during the network's forward propagation. Half-channel $3\times3$ convolutions are applied to all branches, followed by batch normalization (BN)\cite{bn} and activation function layers. Subsequently, a $1\times1$ convolution layer is used to compress the output feature channels to match the input features channels. The concatenation of half-channel 3×3 convolutions from all layers, followed by the passage through the batch normalization layer and activation function, constitutes the process termed as \textbf{Optimizing Channel Feature Jointly (OCFJ)}. OCFJ concatenates channel features generated by multiple branches into a unified distribution, and then refines channel weights using adaptable learnable parameters, $\gamma$ and $\beta$. This method equalizes the weighting of channel features across the branches, enhancing prominent features while suppressing suboptimal ones. If each branch produces distinctive channel feature distributions and employs weighted methods, the process of feature selection becomes prone to being trapped in local optima, thereby hindering the enhancement of network performance. Nevertheless, OCFJ has the capacity to globally filter out high-quality features from among distinct branch feature sets, thereby mitigating the impact of suboptimal features on network performance and substantially reducing both parameters and GFLOPs, as demonstrated in the comparison between the OCFJ and $3\times3$ structures. To comprehensively evaluate the merits of our proposed OCFJ approach, we conducted experiments(Tab.\ref{tab:comp}) using DF (OCFJ) and DF ($3\times3$, excluding OCFJ). Empirical results indicate a notable 0.3\% increment in AP upon OCFJ implementation, accompanied by reductions of 4.5M parameters and 8.2 GFLOPs. This conclusive evidence robustly underscores the inherent advantages of OCFJ.

\begin{table}[h]
\centering
	\caption{The AP(\%), Parameters, GFLOPs of CSPVoVNet, ELAN, and DF, using YOLOv5l-R6.1 as the baseline, on the MS-COCO val2017. In the table, green indicates the best AP.}
	\label{tab:backbone}
	\begin{tabular}{llll}
	\toprule
    Models & AP(\%) & Params(M) & GFLOPs\\
    \midrule
    v5l & 49.0 & 46.5 & 109.1 \\
    v5l+CSPVoVNet & 49.6 (+0.6)& 62.7 (+16.2) & 167.4 (+58.3) \\
    v5l+ELAN & 49.0 & 59.9 (+13.4) & 159.2 (+50.1) \\
    v5l+DF & 49.7 (\textcolor{green}{+0.7}) & 52.4 (+5.9) & 113.1 (+4.0) \\
    \bottomrule
\end{tabular}
\end{table}

To better analyze the advantages of our proposed the Dense Channel Compression for Feature Spatial Solidification structure (DF), we utilize the YOLOv5l-R6.1\cite{yolo5} model as the baseline and replace the original CSP module with the CSPVoVNet, ELAN, and DF modules, respectively. We then perform a comparative analysis on the MS-COCO val2017\cite{mscoco}, focusing on the dimensions of AP(\%), Parameters, and GFLOPs. As evidenced in Tab.\ref{tab:backbone}, our proposed DF module not only substantially reduces the number of parameters and GFLOPs compared to CSPVoVNet and ELAN but also outperforms all comparison models in terms of AP(\%).

\subsection{Asymmetric Multi-level Channel Compression Decoupled Head (ADH)}

The decoupled head module in YOLOX is the first instance of a decoupled head structure employed within the YOLO algorithm series. This structure primarily aims to resolve the conflict between regression and classification tasks. Although YOLOX's decoupled head effectively enhances the model's detection accuracy, it is accompanied by several non-negligible issues. Firstly, this structure considerably increases the number of parameters and GFLOPs, leading to a reduction in inference speed. This reduction contradicts the optimization objectives of real-time object detection models. Secondly, the structure demands a higher consumption of device memory, rendering its application on memory-limited devices, such as edge computing or mobile devices, challenging. Lastly, the decoupled head structure enforces the compression of the feature channel count of input feature maps from all scale detection heads to 256, which leads to a certain degree of feature loss. This limitation prevents it from fully utilizing the high-level features extracted by the backbone and neck networks, resulting in suboptimal model performance, particularly when addressing complex tasks.


We have conducted a series of research and experiments to address the issue of the decoupled head in the YOLOX model. Our findings reveal a logical correlation between the utilization of the decoupled head structure and the associated loss function. Specifically, for different tasks, the structure of the decoupled head should be adjusted in accordance with the complexity of the loss calculation. Furthermore, when applying the decoupled head structure to various tasks, a direct compression of feature channels from the preceding layer into task channels leads to significant feature loss due to differences in the final output dimensions. This, in turn, adversely impacts the overall performance of the model.


 To address both of these issues, we will first analyze the YOLOv5's loss function, which can be summarized as follows:
\begin{equation}
\mathcal{L}_{CIoU}(b_{positive},b_{gt}) = 1-CIoU(b_{positive},b_{gt})
\end{equation}
\begin{equation}
\mathcal{L}_{cls}(c_{positive},c_{gt}) = BCE_{cls}^{sig}(c_{positive},c_{gt};w_{cls})
\end{equation}
\begin{equation}
\mathcal{L}_{obj}(p_{all},p_{iou}) = BCE_{obj}^{sig}(p_{all},p_{iou};w_{obj})
\end{equation}
where $\mathcal{L}_{CIoU}$ represents the Complete Intersection over Union (CIoU)\cite{ciou} loss of bounding boxes. $b_{positive}$ denotes the predicted bounding box values after positive sample matching, while $b_{gt}$ refers to the ground truth of these bounding boxes. Additionally, $\mathcal{L}_{cls}$ is the binary cross-entropy loss for classification, with $c_{positive}$ as the predicted classification values after positive sample matching and $c_{gt}$ as the corresponding ground truth values. Furthermore, $\mathcal{L}_{obj}$ signifies the objectness score for all positive and negative samples. $p_{all}$ represents the predicted objectness scores for both positive and negative samples, whereas $p_{iou}$ encompasses the ground truth objectness scores CIoU values for all positive samples\cite{balanced} and the objectness scores of 0 for negative samples.

In YOLOv5, the total loss is computed as the sum of three components: objectness score loss, classification loss, and bounding box regression loss, which are then used for backpropagation to update the model weights. Since the objectness score is the key parameter for determining whether the predicted grid contains the object, the objectness score loss is calculated based on the difference between the objectness score of the predicted grid and the ground truth (GT) matched grid as the positive sample loss. Simultaneously, the objectness score losses of unmatched grids are treated as negative sample losses to suppress false positives. This combination of positive and negative sample losses leads to an overrepresentation of objectness score loss in the total loss. Both classification loss and bounding box loss are computed only for positive samples, without the need to suppress the predictions of negative samples for these components. As a result, their contributions to the total loss are relatively smaller. During backpropagation, there is no inherent imbalance in how the total loss propagates through the network structure, as the entire loss is backpropagated uniformly. However, due to the disproportionately large share of objectness score loss in the total loss—and since the objectness score is a critical factor in determining prediction accuracy—there is a need to increase the depth of the network's objectness score prediction branch. This adjustment helps balance the overdominance of objectness score loss in the total loss and improves the accuracy of objectness score predictions, thereby enhancing the overall average precision (AP) of the network.

\begin{figure}
  \centering
  \includegraphics[scale=0.3]{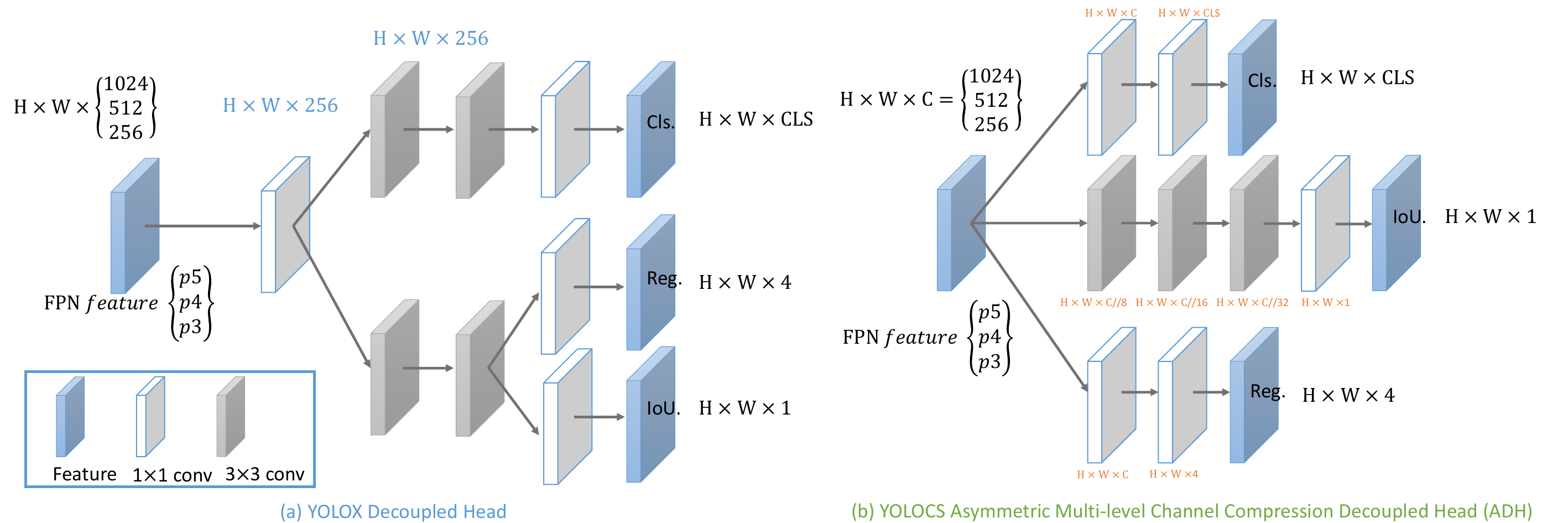}
  \caption{Comparison between YOLOCS's Asymmetric Multi-level Channel Compression Decoupled Head and YOLOX's Decoupled Head}
  \label{fig:adh}
\end{figure}

Additionally, when considering our proposed method of the Dense Channel Compression for Feature Spatial Solidification, directly reducing the number of channels in the final layer to match the output channels can lead to a loss of features during forward propagation, consequently decreasing network performance. Simultaneously, in the context of backpropagation, this structure contributes to suboptimal error backpropagation, impeding the achievement of gradient stability. To address these challenges, we introduce a novel decoupled head called the Asymmetric Multi-level Channel Compression Decoupled Head (Fig.\ref{fig:adh} (b)).


In our proposed ADH, we partition the network based on various task types, completing the corresponding tasks through three distinct network paths (Fig.\ref{fig:adh} (b)). Specifically, we deepen the network path dedicated to the objectness scoring task and employ $3\times3$ convolutions to expand both the receptive field and the number of parameters for this task. Concurrently, we compress the features of each convolutional layer along the channel dimension. This method not only effectively mitigates the training difficulty associated with the objectness scoring task, enhancing model performance, but also substantially reduces the parameters and GFLOPs of the decoupled head (DH) module in YOLOv5, thereby significantly improving inference speed. In addition, we utilized a $1\times1$ convolution to separate the classification and bounding box tasks. This is because the losses associated with both tasks are relatively minor for matched positive samples, and as a result, excessive expansion is avoided. This method substantially decreases the parameters and GFLOPs of the decoupled head (DH) module in YOLOv5, ultimately leading to faster inference speeds.

In order to better assess the benefits of our proposed ADH, we take the YOLOv5l model as our baseline and separately integrate the DH\cite{yolox} from YOLOX and the ADH from YOLOCS. We then conduct a comparative analysis on the MS-COCO val2017, focusing on AP(\%), parameters, and GFLOPs. As evident in Tab.\ref{tab:adh}, our proposed ADH module not only boasts a significantly lower parameters and GFLOPs compared to YOLOX's DH, but also maintains a minimal gap in AP. Consequently, our proposed ADH structure can be considered both innovative and effective.

\begin{table}[h]
\centering
	\caption{The AP(\%), Parameters, GFLOPs for DH and ADH, using YOLOv5l-R6.1 as the baseline, on the MS-COCO val2017. In the table, green indicates the best AP.}
	\label{tab:adh}
	\begin{tabular}{llll}
	\toprule
    Models & AP(\%) & Params(M) & GFLOPs\\
    \midrule
    v5l & 49.0 & 46.5 & 109.1 \\
    v5l+DH & 49.5 (\textcolor{green}{+0.5}) & 53.8 (+7.3) & 150.1 (+41.0) \\
    v5l+ADH & 49.4 (+0.4) & 50.9 (+4.4) & 117.4 (+8.3) \\
    \bottomrule
\end{tabular}
\end{table}


\section{Experiments}

In order to ensure fairness, our experimental setup is maintained in complete alignment with YOLOv5-R6.1. Throughout the training process, we solely integrate our modules, DF and ADH, into the backbone and head networks. All other aspects, including image preprocessing(mosaic\cite{yolo4}, mixup\cite{mixup}, copy-paste\cite{copypaste}), backbone network, neck network, head network, NMS, hyperparameters\footnote{https://github.com/ultralytics/yolov5/blob/master/data/hyps/\\hyp.scratch-high.yaml}, mixed-precision training\cite{mpt}, and random seeds, are kept consistent with YOLOv5. In the inference phase, we adhere to YOLOv5's approach in terms of NMS, rectangular inference, and mixed-precision inference. Our experiments are conducted exclusively on the MS-COCO 2017\cite{mscoco} dataset (MS-COCO 2017 is licensed under a Creative Commons Attribution 4.0 License), utilizing YOLOv5's experimental data, including YOLOX, as baseline for comparison. Each of our models is trained for 300 epochs using the RTX 3090 graphics card.

\subsection{Ablation Experiment}

In our ablation study, we incorporated the two innovative modules, DF and ADH, individually into the YOLOv5l-R6.1 model for comparative analysis on the MS-COCO val2017. Moreover, we combined both modules simultaneously within the YOLOv5l model to further evaluate their performance. For fairness, we incorporated each module subsequently trained for 300 epochs, with the same structures, hyperparameters, and settings. Following the training, we evaluated the models using identical hyperparameters and settings for inference, ultimately obtaining comparative results.

\begin{table}[h]
\centering
	\caption{Ablation Experiment on MS-COCO val2017.}
	\label{tab:ablation}
	\begin{tabular}{ccclll}
	\toprule
    Baseline & DF & ADH & AP(\%) & AP$_{50}$ & AP$_{75}$ \\
    \midrule
    \checkmark  & $\times$ & $\times$ & 49.0 & 67.3 & -  \\
    \checkmark  & \checkmark & $\times$ & 49.7 (\textcolor{green}{+0.7}) & 68.0  & 53.6  \\
    \checkmark  & $\times$ & \checkmark & 49.4 (\textcolor{green}{+0.4}) & 67.4 & 53.9  \\
    \checkmark  & \checkmark & \checkmark & 50.1 (\textcolor{green}{+1.1}) & 67.9 & 54.3  \\
    \bottomrule
\end{tabular}
\end{table}

The data obtained from the ablation experiments, as presented in Tab.\ref{tab:ablation}, reveals that the incorporation of DF and ADH into the YOLOv5 model led to a significant increase of AP by 0.7\% and 0.4\%, respectively. This observation suggest that both of these innovative modules are effective and exhibit satisfactory performance when integrated independently into the model. Moreover, when DF and ADH modules were simultaneously integrated into the YOLOv5 model, the model's AP increased by 1.1\%, indicating that the performance of these two modules is remarkably stable.

\subsection{Comparative Experiment}

\begin{table}
\centering
	\caption{Comparison of YOLOCS, YOLOX and YOLOv5-r6.1\cite{yolo5} in terms of AP on MS-COCO 2017 test-dev\cite{mscoco}. The green highlights indicate the percentage increase in AP achieved by YOLOCS compared to YOLOv5.}
	\label{tab:comparison}
	\begin{tabular}{lllll}
	\toprule
    Models & AP(\%) & Params(M) & GFLOPs & Latency(ms)\\
    \midrule
    v5-S & 37.4 & 7.2 & 16.5 & 6.4 \\
    X-S & 39.6 & 9.0 & 26.8 & 9.8 \\
    CS-S & 42.6 (\textcolor{green}{+5.2}) & 10.6 & 22.9 & 7.2 \\
    \hline
    v5-M & 45.4 & 21.2 & 49.0 & 8.2 \\
    X-M & 46.4 & 25.3 & 73.8 & 12.3 \\
    CS-M & 47.7 (\textcolor{green}{+2.3}) & 29.9 & 64.1 & 8.7 \\
    \hline
    v5-L & 49.0 & 46.5 & 109.1 & 10.1 \\
    X-L & 50.0 & 54.2 & 155.6 & 14.5 \\
    CS-L & 50.1 (\textcolor{green}{+1.1}) & 56.8 & 121.2 & 11.1 \\
    \bottomrule
\end{tabular}
\end{table}
In the comparative experiments, we employed the width (number of channels) and depth (DF depth) of the YOLOCS model as the basis for resizing the model. Specifically, we instantiated YOLOCS as a large, medium, and small model, respectively, following the size configuration approach employed by YOLOv5 and YOLOX. To ensure fairness, we trained models of different sizes for 300 epochs, while maintaining consistency in structures, hyperparameters, and settings. Ultimately, we evaluated the trained models using identical hyperparameters and settings to generate the final comparative results.

\begin{table}
\centering
	\caption{Comparison of the FPS(v100) and AP(\%) of different object detectors on MS-COCO 2017 test-dev\cite{mscoco}. We select all the models
trained on 300 epochs for fair comparison.}
	\label{tab:sota}
	\begin{tabular}{llccc}
	\toprule
    Method & Backbone & Size & FPS & AP(\%)  \\
    \midrule
    RetinaNet  & ResNet-50 & 640 & 37 &  37 \\
	RetinaNet  & ResNet-101 & 640 & 29.4 &  37.9  \\
	RetinaNet  & ResNet-50 & 1024 & 19.6 &  40.1  \\
	RetinaNet  & ResNet-101 & 1024 & 15.4 &  41.1  \\
	\hline
	YOLOv3 + ASFF* & Darknet-53 & 320 & 60 &  38.1  \\
	YOLOv3 + ASFF* & Darknet-53 & 416 & 54 &  40.6  \\
	YOLOv3 + ASFF* & Darknet-53 & 608 & 45.5 &  42.4 \\
	YOLOv3 + ASFF* & Darknet-53 & 800 & 29.4 &  43.9  \\
	\hline
	EfficientDet-D0 & Efficient-B0 & 512 & 62.5 & 33.8  \\
	EfficientDet-D1 & Efficient-B1 & 640 & 50.0 & 39.6  \\
	EfficientDet-D2 & Efficient-B2 & 768 & 41.7 & 43.0  \\
	EfficientDet-D3 & Efficient-B3 & 896 & 23.8 & 45.8  \\
	\hline
	YOLOv4 & CSPDarknet-53 & 608 & 62.0 & 43.5  \\
	YOLOv4-CSP & Modified CSP & 640 & 73.0 & 47.5  \\
	\hline
	YOLOv3-ultralytics & Darknet-53 & 640 & 95.2 & 44.3  \\
	YOLOv5-r6.1-S & Modified CSP v5 & 640 & 156 & 37.4  \\
	YOLOv5-r6.1-M & Modified CSP v5 & 640 & 122 & 45.5\\
	YOLOv5-r6.1-L & Modified CSP v5 & 640 & 99 & 49.0  \\
	\hline
	YOLOX-M & Modified CSP v5 & 640 & 81.3 & 46.4  \\
	YOLOX-L & Modified CSP v5 & 640 & 69.0 & 50.0  \\
	\hline
	YOLOCS-S & DF & 640 & 139 & \textbf{42.6}  \\
	YOLOCS-M & DF & 640 & 115 & \textbf{47.7}  \\
	YOLOCS-L & DF & 640 & 90 & \textbf{50.1}  \\
    \bottomrule
\end{tabular}
\end{table}

Based on the data from the comparative experiment presented in Tab.\ref{tab:comparison}, it can be observed that YOLOCS's small model achieves an AP improvement of 3.0\% and 5.2\% over YOLOX and YOLOv5, respectively. Similarly, YOLOCS's medium model achieves an AP improvement of 1.3\% and 2.3\%, and the large model obtains an improvement of 0.1\% and 1.1\% compared to YOLOX and YOLOv5, respectively. It is worth noting that our proposed YOLOCS outperforms YOLOX and YOLOv5 across all model sizes.

\subsection{Comparison with SOTA}

In order to fully demonstrate the novelty and effectiveness of our proposed YOLOCS model, we compared it with several state-of-the-art object detectors using result data on MS-COCO 2017 test-dev. However, due to the significant impact of software and hardware on the model's inference speed, we calculated the FPS based on the fastest inference speed obtained from multiple attempts. Moreover, for the sake of fairness, we also excluded the latency time associated with image preprocessing and NMS during the inference speed calculation.

As shown in Tab.\ref{tab:sota}, our YOLOCS model surpasses all state-of-the-art detectors with different sizes. Therefore, we propose that the YOLOCS model is both novel and effective, achieving state-of-the-art performance.


\section{Conclusion}
  
In this paper, we address the existing issues in backbone networks and head networks of YOLO by introducing two innovative modules: the Dense Channel Compression for Feature Spatial Solidification Structure and the Asymmetric Multi-level Channel Compression Decoupled Head. The central techniques in these two modules are dense channel compression for feature spatial solidification, which effectively tackle the propagation of features and errors during both forward and backward propagation processes. This enhances the purity of forward-propagated features and strengthens the gradient flow in backward propagation. Simultaneously, although this approach results in a slight increase in the model's parameters and GFLOPs, it still leads to high-level performance. Supported by experimental data, we assert that our proposed YOLOCS achieves state-of-the-art performance.



\section{Acknowledgments}
This work was supported by the National Natural Science Foundation of China [Nos. 62331008, 62027827, 62221005 and 62276040], 
Natural Science Foundation of Chongqing (Nos. 2023NSCQ-LZX0045 and CSTB2022NSCQ-MSX0436)





\bibliographystyle{elsarticle-harv} 
\bibliography{yolocs-clean.bib}



\end{document}